# A possibilistic handling of partially ordered information

**Salem Benferhat**
CRIL - CNRS
Université d'Artois,
Rue Jean Souvraz, SP 18
62307 Lens Cedex, France
benferhat@cril.univ-artois.fr

**Sylvain Lagrue**
SOMEI/PRAXITEC/SIS
115, rue St Jacques,
13006 Marseille, France
lagrue@univ-tln.fr

**Odile Papini**
SIS
Université de Toulon et du Var,
Avenue de l'Université BP 132,
83957 La Garde cedex, France
papini@univ-tln.fr

## Abstract

In standard possibilistic logic, prioritized information are encoded by means of weighted knowledge bases. This paper proposes an extension of possibilistic logic to deal with partially ordered information which can be viewed as a family of possibilistic knowledge bases.

We show that all basic notions of possibilistic logic have natural counterparts when dealing with partially ordered information. Furthermore, we propose an algorithm which computes plausible conclusions of a partially ordered knowledge base.

## 1 Introduction

Possibilistic logic is a weighted logic where each classical logic formula is associated with a weight in $[0, 1]$ understood as a lower bound of a necessity measure. This weight accounts for the level of certainty (or the priority) of the pieces of information represented by the logical formulas.

The unit interval $[0, 1]$ can be understood as a mere ordinal scale. Namely, possibilistic logic is appropriate for reasoning with prioritized information, when priorities are represented by a total pre-order.

There were several attempts for extending possibilistic logic for dealing with other, more complex, uncertainty structures instead of total pre-orders. For instance, in [Dubois et al., 1992, Lafage et al., 1999, de Cooman, 1996], extensions based on lattices have been proposed. The extension proposed in [Dubois et al., 1992] makes sense for representing temporal information, however it does not fully extend the possibilistic logic inference for inconsistent sets of beliefs.

The aim of this paper is to propose a natural extension of possibilistic logic to deal with a partially-ordered knowledge base. It expands and completes results obtained in [Benferhat et al., 2003].

More precisely, we first propose a natural definition of the possibilistic logic inference based on the family of totally ordered knowledge bases (resp. possibility distributions), which are compatible with a partial knowledge base. We then provide a semantic (resp. syntactic) characterization of this inference, which is based on a strict partial order on interpretations (resp. on a strict partial order on consistent subbases of the knowledge base). We then show that the main properties of the possibilistic logic (subsumed formulas, clausal form, soundness and completeness results) hold for a partially ordered knowledge base. Finally, we propose an algorithm for computing the set of plausible conclusions of a partially ordered knowledge base.

The following section gives some basic definitions and a brief refresher on possibilistic logic.

## 2 Basic definitions and background

### 2.1 Partial orders

A partial pre-order $\preceq$ on a *finite* set $A$ is a reflexive ($a \preceq a$) and transitive (if $a \preceq b$ and $b \preceq c$ then $a \preceq c$) binary relation. In this paper, $a \preceq b$ intuitively means that $a$ is at least as preferred as $b$.

A strict partial order $\prec$ on a set $A$ is an irreflexive ($a \prec a$ does not hold) and transitive binary relation. $a \prec b$ means that $a$ is strictly preferred to $b$. A strict partial order is generally defined from a partial pre-order as $a \prec b$ if $a \preceq b$ holds but $b \preceq a$ does not hold.

The equality is defined by $a \approx b$ iff $a \preceq b$ and $b \preceq a$. $a \approx b$ means that $a$ and $b$ are equally preferred. We lastly define incomparability, denoted by $\sim$, as $a \sim b$ if and only if neither $a \preceq b$ nor $b \preceq a$ holds. $a \sim b$ means that neither $a$ is preferred to $b$, nor the converse.

In the following, $a \not\preceq b$ (resp. $a \not\prec b$, $a \neq b$) means that $a \preceq b$ (resp. $a \prec b$, $a = b$) does not hold.

A total pre-order $\leq$ is a partial pre-order such that $\forall a, b \in A : a \leq b$ or $b \leq a$.

Let $\prec$ be a strict partial order on a set $A$. The set of minimal elements of $A$, denoted by $Min(A, \prec)$, is defined as follows: $Min(A, \prec) = \{a : a \in A, \nexists b \in A, b \prec a\}$.

Note that only the strict partial order is useful for determining minimal elements of $A$.

### 2.2 Qualitative possibilistic logic

We only provide a brief background on qualitative possibilistic logic (for more details, see [Dubois et al., 1994, Dubois and Prade, 1998]).

Let $\mathcal{L}$ be a *finite* propositional language. We denote by $\Omega$ the set of interpretations of $\mathcal{L}$, and by $\omega$ an element of $\Omega$. Let $\varphi$ be a formula, $Mod(\varphi)$ denotes the set of models of $\varphi$.

The basic element of the semantics of possibilistic logic is the notion of a possibility distribution [Dubois and Prade, 1988], denoted by $\pi$, which is a function from $\Omega$ to $[0, 1]$. $\pi(\omega)$ evaluates to what extent $\omega$ is compatible, or consistent, with our available knowledge. $\pi(\omega) = 0$ means that $\omega$ is impossible, while $\pi(\omega) = 1$ means that $\omega$ is totally possible. $\pi(\omega) \geq \pi(\omega')$ means that $\omega$ is more plausible than $\omega'$.

At the syntactic level, uncertain pieces of information are represented by means of a set of weighted formulas of the form $K = \{(\varphi_i, a_i) : i = 1, \ldots, n\}$ where $\varphi_i$ is a propositional formula, and $a_i \in ]0, 1]$. The real number $a_i$ represents a lower bound of certainty degree of the formula $\varphi_i$.

Each possibilistic knowledge base induces a unique possibility distribution, denoted by $\pi_\Sigma$, where interpretations are ordered with respect to the highest formula that they falsify [Dubois et al., 1994]. More formally, $\forall \omega \in \Omega$:

$$\pi_\Sigma(\omega) = \begin{cases} 1 \text{ if } \forall(\varphi, a) \in \Sigma : \omega \models \varphi, \\ 1 - max\{a : (\varphi, a) \in \Sigma \text{ and } \omega \not\models \varphi\} \\ \text{otherwise.} \end{cases}$$

Note that all tautologies can be removed from $K$, without changing the induced possibility distribution. Similarly, it has been shown in [Dubois et al., 1994] that subsumed formulas can also be removed. A formula $\varphi$ is said to be subsumed if it is entailed from formulas having a weight strictly greater than $\varphi$.

Lastly, any formulas can equivalently be transformed into its CNF form. If $(\varphi, a)$ belongs to $K$, it can then be replaced by $(C1, a)$, $(C2, a)$, ..., $(C_n, a)$, where $\{C_1, C_2, \ldots, C_n\}$ represents the clausal form of $\varphi$, without any change in the induced possibility distribution [Dubois et al., 1994].

In qualitative possibilistic logic, the interval $[0, 1]$ is simply interpreted as a mere totally ordered scale. In this case, a total pre-order on interpretations, denoted by $\leq_\pi$, can be associated with a possibility distribution $\pi$ in the following way:

$$\forall \omega, \omega' \in \Omega, \omega \leq_\Omega \omega' \text{ iff } \pi(\omega) \geq \pi(\omega').$$

The pair $(\Omega, \leq_\Omega)$ will be called *a qualitative possibilistic distribution*.

A formula $\psi$ is said to be a plausible consequence of $(\Omega, \leq_\Omega)$, denoted by $(\Omega, \leq_\Omega) \models_\pi \psi$, iff $Min(\Omega, <_\Omega) \subseteq Mod(\psi)$.

A possibilistic knowledge base can also be represented qualitatively by a pair $(\Sigma, \leq_\Sigma)$, where $\Sigma$ is a set of formulas and $\leq_\Sigma$ is a total pre-order on $\Sigma$, in the following way (if $\Sigma$ contains two syntacticly identical formulas, one of them is then replaced by another syntacticly different, but logically equivalent formula):

$$\varphi \leq_\Sigma \varphi' \text{ iff } (\varphi, a), (\varphi', b) \in \Sigma \text{ and } b \leq a.$$

A pair $(\Sigma, \leq_\Sigma)$ will be called a *totally ordered knowledge base*. A syntactic inference, denoted by $(\Sigma, \leq_\Sigma) \vdash_\pi \psi$, where $\psi$ represents an inferred formula, can be achieved with a computational complexity slightly higher than the one of classical logic (see [Lang, 2001] for more details).

Similarly, each totally ordered knowledge base $(\Sigma, \leq_\Sigma)$ induces a unique qualitative distribution $(\Omega, \leq_\Omega)$ defined by:

**(eqn i)** $\omega \leq_\Omega \omega'$ if $\forall \varphi \in \Sigma$ such that $\omega \not\models \varphi$, $\exists \psi \in \Sigma$ such that $\omega' \not\models \psi$ and $\psi \leq \varphi$.

The syntactic inference coincides with the semantic one. Namely, $\forall \psi \in \mathcal{L}, (\Omega, \leq_\Omega) \models_\pi \psi$ iff $(\Sigma, \leq_\Sigma) \vdash_\pi \psi$.

## 3 Representing partially ordered information

Generally, an agent cannot provide a total pre-order between all of his beliefs, but only a partial pre-order. A partially ordered knowledge base is a pair $(\Sigma, \preceq_\Sigma)$ where $\Sigma$ is a set of propositional formulas and $\preceq_\Sigma$ is a partial pre-order (i.e. a transitive and reflexive relation) on $\Sigma$.

In the following, we will use the following example for illustrating different notions used in this paper.

**Example 1** *Let $(\Sigma, \preceq_\Sigma)$ be a partially ordered knowledge base such that $\Sigma = \{\neg a, a, \neg b, b\}$ and $\preceq_\Sigma$ is the partial pre-order on $\Sigma$ represented by Figure 1. The arrow $\neg a \to b$ means that $b \preceq_\Sigma \neg a$ (for the sake of simplicity, neither transitivity nor reflexivity are represented).*

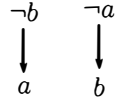

Figure 1: Example of $\preceq_\Sigma$

*Namely, $\preceq_\Sigma$ is such that: $a \prec_\Sigma \neg b$ and $b \prec_\Sigma \neg a$. Note that $\Sigma$ is inconsistent.*

The natural question is how to define an inference relation for $(\Sigma, \preceq_\Sigma)$, which extends the possibilistic logic inference when $\preceq_\Sigma$ is a total pre-order.

### 3.1 Encoding partial pre-order by means of boolean lattice

In their attempt to cast both uncertainty and time in a logical framework, Dubois and al. use a boolean lattice $S^T$, where $T$ represents a set of possible times, instead of the interval $[0, 1]$ [Dubois et al., 1992] (see also [Lafage et al., 1999] for similar works based on De Kleer's ATMS [de Kleer, 1986]). $\pi(\omega) = \mathcal{T} \subseteq T$ means that at any instantiation in $\mathcal{T}$, $\omega$ is possible (not excluded). Necessity and possibility are naturally extended where roughly speaking the maximum, minimum and reversing scale $(1 - (.))$ are replaced by the union, intersection and complementary. A weighted formula has the form $(\varphi, \mathcal{T})$, with $\mathcal{T} \subseteq T$, which means that $\varphi$ is true at least during the set of time intervals $\mathcal{T}$. All other basic concepts of possibilistic logic have natural counterparts when using boolean lattice instead of the unit interval $[0, 1]$.

The possibilistic logic machinery is also extended. First the resolution rule proposed is:

$$(C, \tau), (C', \tau') \vdash (C \wedge C', \tau \cap \tau')$$

Then, given $F$ a set of weighted formulas, we define $Inc$ as $Inc(F) = \bigcup\{\tau, F \vdash (\bot, \tau)\}$. Lastly, $\psi$ is a plausible consequence of $F$ if $Inc(F \cup \{(\neg\psi, T)\}) > Inc(F)$.

This extension completely makes sense to handle temporal information in the possibilistic logic setting. However, the proposed extension is not fully satisfying for inconsistent beliefs.

**Example 2** *For instance, let us consider example 1, where $T = \{t_1, t_2, t_3\}$ is a set of 3 time instants. The representation of our example using the valuation lattice of Figure 2 can be:*

$$F = \{(a, \{t_1, t_2\}), (b, \{t_2, t_3\}), (\neg b, \{t_1\}), (\neg a, \{t_3\})\}.$$

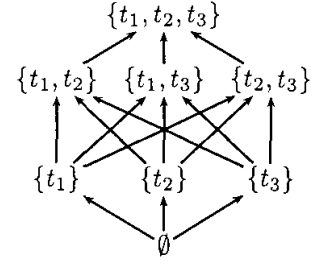

Figure 2: A boolean lattice

*It can be checked that $Inc = \bigcup\{\tau, F \vdash (\bot, \tau)\} = \emptyset$. Indeed, there are two ways for reaching $\bot$: $(a, \{t_1, t_2\}), (\neg a, \{t_3\}) \vdash (\bot, \emptyset)$ and $(b, \{t_2, t_3\}), (\neg b, \{t_1\}) \vdash (\bot, \emptyset)$. Lastly, it can be checked that both $a$ and $\neg a$ are plausible consequence (for instance, $Inc(F \cup \{(a, \{t_1, t_2, t_3\})\}) > \emptyset$), which is not desirable.*

### 3.2 Compatible possibilistic knowledge bases

A natural way for representing $(\Sigma, \preceq_\Sigma)$ is to consider the set of all compatible totally ordered knowledge bases $(\Sigma, \leq_\Sigma)$. Intuitively, $(\Sigma, \leq_\Sigma)$ is said to be compatible with $(\Sigma, \preceq_\Sigma)$ if $\leq_\Sigma$ extends $\preceq_\Sigma$, namely it preserves strict and non-strict preference relations between any two formulas of $\Sigma$. More formally:

**Definition 1** *Let $(\Sigma, \preceq_\Sigma)$ be a partially ordered base and $\prec_\Sigma$ its usual strict counterpart. A totally ordered knowledge base $(\Sigma, \leq_\Sigma)$ is then said to be compatible with $(\Sigma, \preceq_\Sigma)$ iff:*

*(i) $\forall \varphi, \varphi' \in \Sigma$: if $\varphi \prec_\Sigma \varphi'$ then $\varphi <_\Sigma \varphi'$,*

*(ii) $\forall \varphi, \varphi' \in \Sigma$: if $\varphi \preceq_\Sigma \varphi'$ then $\varphi \leq_\Sigma \varphi'$.*

The set of all totally ordered knowledge bases compatible with $(\Sigma, \preceq_\Sigma)$ is denoted by $\mathcal{F}(\Sigma, \preceq_\Sigma)$.

**Example 3** *Let us consider again the partially ordered knowledge base $(\Sigma, \preceq_\Sigma)$ defined in example 1. Two examples of a totally ordered knowledge base compatible with $(\Sigma, \preceq_\Sigma)$ are $(\Sigma, \leq_{1,\Sigma})$ and $(\Sigma, \leq_{2,\Sigma})$, defined by:*

$$a \approx_{1,\Sigma} b <_{1,\Sigma} \neg a <_{1,\Sigma} \neg b$$
$$a <_{2,\Sigma} \neg b <_{2,\Sigma} b <_{2,\Sigma} \neg a.$$

Given $\mathcal{F}(\Sigma, \preceq_\Sigma)$, it is now possible to define a plausible inference relation from a partially order knowledge base. A formula $\varphi$ is a plausible consequence of a partially ordered knowledge base $(\Sigma, \preceq_\Sigma)$ if it is a possibilistic conclusion for each compatible possibilistic knowledge base. More formally:

**Definition 2** *A formula $\psi$ is said to be a plausible consequence of $(\Sigma, \preceq_\Sigma)$, denoted by $(\Sigma, \preceq_\Sigma) \vdash_{po} \psi$, if $\psi$ is a possibilistic consequence of each totally ordered knowledge base compatible with $(\Sigma, \preceq_\Sigma)$, namely:*

$$(\Sigma, \preceq_\Sigma) \vdash_{po} \psi \text{ iff } \forall (\Sigma, \leq_\Sigma) \in \mathcal{F}(\Sigma, \preceq_\Sigma), (\Sigma, \leq_\Sigma) \vdash_\pi \psi.$$

Note that this syntactic definition is equivalent to the following semantic definition:

**Definition 3** *Let $\mathcal{F}(\Sigma, \preceq_\Sigma)$ be a set of all compatible totally ordered knowledge bases. Let $\mathcal{F}(\Omega, \preceq_\Omega)$ be the family of all qualitative possibility distributions induced from each element of $\mathcal{F}(\Sigma, \preceq_\Sigma)$ using **(eqn i)**. Then:*

$(\Omega, \preceq_\Omega) \models_{po} \psi$ *iff*

$$\forall (\Omega, \leq_\Omega) \in \mathcal{F}(\Omega, \preceq_\Omega), (\Omega, \leq_\Omega) \models_\pi \psi.$$

**Example 4** *Let us consider the partially ordered knowledge base provided by Example 1. The formula $b$ is not a plausible conclusion of $(\Sigma, \preceq_\Sigma)$. Indeed it cannot be inferred from the compatible totally ordered knowledge base $(\Sigma, \leq_{2,\Sigma})$ which is such that*

$$a <_{2,\Sigma} \neg b <_{2,\Sigma} b <_{2,\Sigma} \neg a.$$

*Similarly $a \wedge b$ is not a plausible conclusion, indeed it cannot be inferred from the compatible totally ordered knowledge base $(\Sigma, \leq_{2,\Sigma})$.*

*However, the formula $a \vee b$ is a plausible conclusion of $(\Sigma, \preceq_\Sigma)$, since neither $\neg a$ nor $\neg b$ can be among the most plausible formulas in any totally ordered knowledge base (otherwise, either $a < \neg b$ or $b < \neg a$ is violated). This means that either $a$ or $b$ is among the most plausible formula in each totally ordered knowledge base. Hence $a \vee b$ is a plausible conclusion of each totally ordered knowledge base.*

Clearly, the number of compatible possibilistic knowledge bases can be very large. The following provides a semantic and syntactic (with a computational issue section) inference from partially ordered knowledge base without using a family of possibility distribution (resp possibility knowledge base).

## 4 A semantic representation for reasoning with partially ordered information

The idea of this section is that, instead of computing all compatible possibility distributions, we define a strict partial order on interpretations which preserves the possibilistic inference provided by definition 2.

Roughly speaking, the interpretation $\omega$ is strictly preferred to the interpretation $\omega'$, denoted by $\omega \lhd_\Omega \omega'$, if the set of formulas falsified by $\omega'$ is preferred to the set of formulas falsified by $\omega$.

Therefore, we need to define a preference relation on subsets of formulas from a partial pre-order on a set of formulas. There were several ways to define such relations (see [Halpern, 1997], [Cayrol et al., 1992] and [Benferhat et al., 2003] for more details). We use the following one:

**Definition 4** *Let $(\Sigma, \preceq_\Sigma)$ be a partially ordered knowledge base and $X, Y \subseteq \Sigma$ (we assume that neither $X$ nor $Y$ is empty), $X$ is strictly preferred to $Y$, denoted by $X \lhd Y$ iff:*

$$\forall y \in Min(Y, \prec), \exists x \in Min(X, \prec) \text{ such that } x \prec y.$$

Note that If $X = Y = \emptyset$, we then consider that neither $X \lhd Y$ holds, nor $Y \lhd X$.

We can now define $\lhd_\Omega$. For this aim, let us denote $\lceil \omega, \Sigma \rceil$ the set of preferred formulas in $\Sigma$ falsified by $\omega$, namely: $\lceil \omega, \Sigma \rceil = \{\varphi : \varphi \in \Sigma, \omega \not\models \varphi\}$.

**Definition 5** *Let $(\Sigma, \preceq_\Sigma)$ be a partially ordered knowledge base, then $\omega$ is said to be strictly preferred to $\omega'$ according to $(\Sigma, \preceq_\Sigma)$, denoted by $\omega \lhd_\Omega \omega'$ iff:*

$$\lceil \omega', \Sigma \rceil \lhd \lceil \omega, \Sigma \rceil.$$

It can be shown that $\lhd_\Omega$ is a strict partial order.

**Example 5** *Let us consider again Example 1. Table 1 shows the set of formulas falsified by each interpretations.*

Table 1: Set of falsified formulas for each interpretations

| $\Omega$ | $a$ | $b$ | $\{\varphi \in \Sigma : \omega_i \not\models \varphi\}$ | $\lceil \omega_i, \Sigma \rceil$ |
|---|---|---|---|---|
| $\omega_0$ | $\neg a$ | $\neg b$ | $\{a, b\}$ | $\{a, b\}$ |
| $\omega_1$ | $\neg a$ | $b$ | $\{a, \neg b\}$ | $\{a\}$ |
| $\omega_2$ | $a$ | $\neg b$ | $\{\neg a, b\}$ | $\{b\}$ |
| $\omega_3$ | $a$ | $b$ | $\{\neg a, \neg b\}$ | $\{\neg a, \neg b\}$ |

*Using the definition of $\lhd$, we can show that $\omega_3 \lhd_\Omega \omega_0$. Indeed $\lceil \omega_3, \Sigma \rceil = \{\neg a, \neg b\}$ and $\lceil \omega_0, \Sigma \rceil = \{a, b\}$. Moreover we have: $b \prec_\Sigma \neg a$ and $a \prec_\Sigma \neg b$, hence $\{a, b\} \lhd \{\neg a, \neg b\}$, which implies, following Definition 4, $\omega_3 \lhd_\Omega \omega_0$.*

The definition of the semantic inference is as follows [Benferhat et al., 2003]:

**Definition 6** *Let $\lhd_\Omega$ be the partial pre-order on $\Omega$ induced from a partially ordered base $(\Sigma, \preceq_\Sigma)$, using definition 5. A formula $\psi$ is inferred from $(\Omega, \lhd_\Omega)$, denoted by $(\Omega, \lhd_\Omega) \models \psi$, iff $Min(\Omega, \lhd_\Omega) \subseteq Mod(\psi)$.*

Let us illustrate this definition by the following example.

**Example 6** *We consider again Example 1.*

*The formula $a \vee b$ can be inferred, indeed we have $Mod(a \vee b) = \{\omega_1, \omega_2, \omega_3\}$, $Min(\Omega, \triangleleft_\Omega) = \{\omega_1, \omega_2, \omega_3\}$ and $Min(\Omega, \triangleleft_\Omega) \subseteq Mod(a \vee b)$. We then have $(\Omega, \triangleleft_\Omega) \models a \vee b$.*

*However, the formula $a \wedge b$ cannot be inferred. Indeed we have $Mod(a \wedge b) = \{\omega_3\}$, which does not contain $Min(\Omega, \triangleleft_\Omega)$.*

The following theorem shows that this definition is equivalent to the inference based on the family of totally ordered knowledge bases proposed in Definition 2:

**Theorem 7** *Let $(\Sigma, \preceq_\Sigma)$ be a partially ordered knowledge base, $(\Omega, \triangleleft_\Omega)$ as given by definition 5 and $\mathcal{F}(\Sigma, \preceq_\Sigma)$ the family of all compatible possibilistic distribution then: $\forall \psi \in \mathcal{L}, (\Sigma, \preceq_\Sigma) \vdash_{po} \psi$ iff $(\Omega, \triangleleft_\Omega) \models \psi$.*

Theorem 7 provides a strong justification of the inference relation given by definition 6 and proposed in [Benferhat et al., 2003]. In this paper, [Benferhat et al., 2003], a second inference relation has been proposed. This inference is based on $\triangleleft'_\Omega$ defined as: $\omega \triangleleft'_\Omega \omega'$ iff $\exists \varphi' \in [\omega', \Sigma]$ such that $\forall \varphi \in [\omega, \Sigma], \varphi' \prec_\Sigma \varphi$.

The inference is defined as $(\triangleleft'_\Omega, \Omega) \vdash \psi$ iff $Min(\Omega, \triangleleft'_\Omega) \subseteq Mod(\psi)$.

This inference violates theorem 7, as it is shown by the following (counter-)example:

**Example 7** *We consider again Example 1, if we compute $\triangleleft'_\Omega$, we obtain that all interpretations are incomparable. Indeed there exists no formula in the example which is preferred to all formulas falsified by another interpretation (each interpretations falsifying at last 2 formulas). We then have $Min(\triangleleft'_\Omega, \Omega) = \Omega$. Only tautologies can be inferred.*

As in possibilistic logic, tautologies can be removed from an initial partially ordered knowledge base, without any change in the plausible conclusions. This is simply due to the fact that the computation of $\triangleleft_\Omega$ is based on falsified formulas, and tautologies are satisfied by all interpretations.

The following definition provides a natural extension of the notion of subsumed formulas.

**Definition 8** *Let $(\Sigma, \preceq_\Sigma)$ be a partially ordered knowledge base. Let $\varphi \in \Sigma$ and $Pref(\varphi, \preceq_\Sigma) = \{\psi \in \Sigma : \psi \preceq_\Sigma \varphi\}$.*

*$\varphi \in \Sigma$ is said to be subsumed by $(\Sigma, \preceq_\Sigma)$ if: $Pref(\varphi, \preceq_\Sigma) \vdash \varphi$.*

As in possibilistic logic, the subsumed formulas can be removed without any change in our inference:

**Proposition 9** *Let $(\Sigma, \preceq_\Sigma)$ be a partially ordered knowledge base and $Sub \subset \Sigma$ be the set. Let $\Sigma'$ be such that $\Sigma' = \Sigma \setminus Sub$ and $\preceq'_\Sigma$ be the restriction of $\Sigma$ to the elements of $\Sigma'$. Then:*

$$\forall \psi, (\Sigma, \preceq_\Sigma) \vdash_{po} \psi \text{ iff } (\Sigma', \preceq'_\Sigma) \vdash_{po} \psi.$$

Again as in possibilistic logic, we can transform any formulas into its CNF form, without any change in the inference.

**Definition 10** *Let $(\Sigma, \preceq_\Sigma)$ be a partially ordered knowledge base. Let $\varphi \in \Sigma$ and $\{C_1, C_2, \ldots, C_n\}$ its clausal form. The partially ordered knowledge base $(\Sigma, \preceq_\Sigma)$ can be translated into the knowledge base $(\Sigma', \preceq_{\Sigma'})$ by the following way: $\Sigma' = (\Sigma \setminus \{\varphi\}) \cup \{C_1, C_2, \ldots, C_n\}$ and $\forall \varphi' \in \Sigma \setminus \{\varphi\}, C_i \in \{C_1, C_2, \ldots, C_n\}$ :*

- *if $\varphi' \preceq_\Sigma \varphi$ (resp. $\varphi' \prec_\Sigma \varphi$) then $\varphi' \preceq_{\Sigma'} C_i$ (resp. $\varphi' \prec_{\Sigma'} \varphi$),*

- *if $\varphi \preceq_\Sigma \varphi'$ (resp $\varphi \prec_\Sigma \varphi'$) then $C_i \preceq_{\Sigma'} \varphi'$ (reps. $C_i \prec_{\Sigma'} \varphi'$),*

- *if $\varphi' \sim_\Sigma \varphi$ then $\varphi' \sim_{\Sigma'} C_i$,*

*and $\forall C_i, C_j \in \{C_1, C_2, \ldots, C_n\} : C_i \approx_{\Sigma'} C_i$.*

**Proposition 11** *Let $(\Sigma, \preceq_\Sigma)$ be a partially ordered knowledge base and $(\Sigma', \preceq_{\Sigma'})$ given by Definition 10. Then: $(\Sigma, \preceq_\Sigma) \vdash_{po} \psi$ iff $(\Sigma', \preceq_{\Sigma'}) \vdash_{po} \psi$.*

Lastly, the following proposition shows that, when $\preceq_\Sigma$ is a total pre-order, our inference relation is an extension of qualitative possibilistic logic:

**Proposition 12** *Let $(\Sigma, \preceq_\Sigma)$ be a partially ordered knowledge base such that $\preceq_\Sigma$ is a total pre-order then:*

$$(\Sigma, \preceq_\Sigma) \vdash_{po} \psi \text{ iff } (\Sigma, \leq_\Sigma) \vdash_\pi \psi.$$

## 5 Syntactic representation

The aim of this section is to provide a syntactic inference directly achieved from $(\Sigma, \preceq_\Sigma)$. For this aim, we define a strict partial order, denoted by $\triangleleft_\mathcal{C}$ between all consistent subsets of $\Sigma$, denoted by $\mathcal{C}$ [Benferhat et al., 2003].

**Definition 13** *Let $C_1, C_2 \in \mathcal{C}$ be two consistent subsets of $\Sigma$, $C_1$ is said to be preferred to $C_2$, denoted by $C_1 \triangleleft_\mathcal{C} C_2$, iff: $\{\varphi_2 \notin C_2\} \triangleleft \{\varphi_1 \notin C_1\}$.*

Intuitively, $C_1$ is preferred to $C_2$ if the preferred elements outside $C_1$ are less important than the preferred elements outside $C_2$. This is a natural extension of $BO$ ("Best Out") ordering used in [Benferhat et al., 1993] for totally ordered information.

**Example 8** *Let us again consider Example 1. $\mathcal{C}$ is composed of 9 consistent subsets, which are listed in Table 2.*

Table 2: Set of consistent subsets of $\Sigma$.

| $C_i$ | $\{\varphi \in C_i\}$ | $\{\varphi \notin C_i\}$ | $Min(\{\varphi \notin C_i\}, \prec_\Sigma)$ |
|---|---|---|---|
| $C_0$ | $\emptyset$ | $\{a, \neg a, b, \neg b\}$ | $\{a, b\}$ |
| $C_1$ | $\{a\}$ | $\{\neg a, b, \neg b\}$ | $\{\neg b, b\}$ |
| $C_2$ | $\{\neg a\}$ | $\{a, b, \neg b\}$ | $\{a, b\}$ |
| $C_3$ | $\{b\}$ | $\{a, \neg a, \neg b\}$ | $\{a, \neg a\}$ |
| $C_4$ | $\{\neg b\}$ | $\{a, \neg a, b\}$ | $\{a, b\}$ |
| $C_5$ | $\{a, b\}$ | $\{\neg a, \neg b\}$ | $\{\neg a, \neg b\}$ |
| $C_6$ | $\{a, \neg b\}$ | $\{\neg a, b\}$ | $\{b\}$ |
| $C_7$ | $\{\neg a, \neg b\}$ | $\{a, b\}$ | $\{a, b\}$ |
| $C_8$ | $\{\neg a, b\}$ | $\{a, \neg b\}$ | $\{a\}$ |

*For instance, we have $C_5 \lhd_\mathcal{C} C_0$. Indeed for each element outside $C_5$ ($\neg a$ or $\neg b$), there exists a strictly preferred element outside $C_0$ (resp. $b$ and $a$). Again $C_5 \lhd_\mathcal{C} C_0$ is the only relation between the elements of $\mathcal{C}$. Similarly, it can be shown that $Min(\mathcal{C}, \lhd_\mathcal{C}) = \{C_1, C_3, C_5, C_6, C_8\}$.*

We denote by $Cons$ the set of preferred consistent subsets of $\Sigma$, with respect to $\lhd_\mathcal{C}$. Namely: $Cons = Min(\mathcal{C}, \lhd_\mathcal{C})$.

We can now define a syntactic inference:

**Definition 14** *A formula $\psi$ is syntacticly inferred from $(\Sigma, \preceq_\Sigma)$, which is denoted by $(\Sigma, \lhd_\mathcal{C}) \vdash \psi$ iff $\forall C \in Cons, C \vdash \psi$.*

**Example 9** *We have $Cons = \{C_1, C_3, C_5, C_6, C_8\}$, with $C_1 = \{a\}$, $C_3 = \{b\}$, $C_5 = \{a, b\}$, $C_6 = \{a, \neg b\}$, $C_8 = \{\neg a, b\}$.*

*Let us show that the formula $\psi = a \vee b$ is a consequence of $(\Sigma, \lhd_\mathcal{C})$. We have $\neg\psi = \neg a \wedge \neg b$ and $C_1 \cup \{\neg a \wedge \neg b\}$ is inconsistent, as well as $C_3 \cup \{\neg a \wedge \neg b\}$, $C_5 \cup \{\neg a \wedge \neg b\}$, $C_6 \cup \{\neg a \wedge \neg b\}$ and $C_8 \cup \{\neg a \wedge \neg b\}$.*

*Therefore $(\Sigma, \lhd_\mathcal{C}) \vdash a \vee b$.*

The following theorem shows that Definition 14 is indeed a syntactic characterization of the $(\Sigma, \preceq_\Sigma) \vdash_{po}$, namely:

**Theorem 15** *Let $(\Sigma, \preceq_\Sigma)$ be a partially ordered knowledge base, then: $\forall \psi \in \mathcal{L} : (\Sigma, \preceq_\Sigma) \vdash_{po} \psi$ iff $(\Sigma, \lhd_\mathcal{C}) \vdash \psi$.*

## 6 Computing syntactic inference

Clearly, the size of $Cons$ can be very large. The following shows that not all elements of $Cons$ are. We provide several properties for reducing the size of $Cons$ and for computing it.

### 6.1 Some properties of $Cons$

The following shows that only minimal elements of $Cons$ for the set inclusion are useful for the inference. The minimal elements of $Cons$ are denoted by $Ker$. More formally:

$$Ker = Min(Cons, \subseteq)$$

Indeed we have:

**Proposition 16** *Let $(\Sigma, \preceq_\Sigma)$ be a partially ordered knowledge base, $\psi$ a formula. Then:*

$$(\Sigma, \preceq_\Sigma) \vdash_{po} \psi \text{ iff } \forall C \in Ker, C \vdash \psi.$$

**Example 10** *In Example 1, we have $Cons = \{\{a\}, \{a, b\}, \{a, \neg b\}, \{b\}, \{\neg a, b\}\}$ from which we can deduce $a \vee b$. Moreover, we have $Ker = \{\{a\}, \{b\}\}$ from which we can also deduce $a \vee b$.*

Restricting to $Ker$ is justified by the fact that if $C \in Ker$ and $C \vdash \psi$, then $\forall C' \in Cons, C \subseteq C', C' \vdash \psi$.

The following defines the notion of a "rooted" consistent subset of $\Sigma$. A consistent subset $C$ of $\Sigma$ is said to be rooted if and only if for each formula $\varphi$ in $C$, all preferred formulas of $\varphi$ are in $C$. More formally:

**Definition 17** *Let $C$ be a consistent subset of $\Sigma$, $\varphi \in C$ and $Pref(\varphi, \preceq_\Sigma) = \{\psi \in \Sigma : \psi \prec \varphi\}$. $C$ is said to be rooted, if and only if:*

$$\forall \varphi \in C, Pref(\varphi, \preceq_\Sigma) \subseteq C.$$

**Example 11** *In our example $C = \{a, \neg b\}$ is a rooted consistent subset. We have $Pref(a, \preceq_\Sigma) = \emptyset$ and $Pref(\neg b, \preceq_\Sigma) = a$. Clearly $Pref(a, \preceq_\Sigma) \subseteq C$ and $Pref(a, \preceq_\Sigma) \subseteq C$*

*However, $C' = \{\neg a\}$ is not a rooted consistent subset. Indeed $Pref(\neg a, \preceq_\Sigma) = \{b\}$ is not in $C'$.*

The collection of all rooted elements of $\mathcal{C}$ is denoted by $Root(\mathcal{C})$.

The following shows that all elements of $Ker$ are "rooted", more formally:

**Proposition 18** *Let $(\Sigma, \preceq_\Sigma)$ be a partially ordered knowledge base, then: $Ker \subseteq Root(\mathcal{C})$.*

The following proposition provides a condition satisfied by all element in $Cons$.

**Proposition 19** *Suppose that $\Sigma$ is inconsistent. Then, $\forall C \in Cons, C \cup Min(\Sigma \setminus C, \prec_{\Sigma \setminus C})$ is inconsistent.*

This proposition means that for each element $C$ which is in $Cons$, the addition of minimal elements of $\Sigma \setminus C$ to $C$ leads to a conflict.

**Example 12** *We have $C_6 = \{a, \neg b\}$ which belongs to $Cons$. We have $Min(\Sigma \setminus C_6, \prec_{\Sigma \setminus C_6}) = \{b\}$ and $\{a, \neg b, b\}$ which is inconsistent.*

The converse of Proposition 19 does not hold. For instance, let us consider $\Sigma = \{a, \neg a\}$ and $a \prec \neg a$. We have then $Min(\Sigma \setminus \{\neg a\}, \prec_{\Sigma \setminus \{\neg a\}}) \cup \{\neg a\} = \{a, \neg a\}$, which is inconsistent. But $\{\neg a\}$ does not belong to $Cons$.

However, the converse holds for the rooted consistent sets.

**Proposition 20** *Let $C \in Root(\mathcal{C})$. Then if $C \cup Min(\Sigma \setminus C, \prec_{\Sigma \setminus C})$ is inconsistent, then $C \in Cons$.*

Figure 3 summarizes the relationships (in term of set inclusion) between the consistent subsets of $\Sigma$, denoted by $\mathcal{C}$, its rooted elements, denoted by $Root(\mathcal{C})$, $Ker$ and $Cons$,

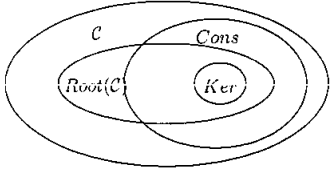

Figure 3: Set inclusion relations between $Cons$, $Ker$, $\mathcal{C}$ and $Root(\mathcal{C})$

### 6.2 An algorithm for computing $Ker$

We now propose an algorithm (Algorithm 1) which uses the properties of the previous section for computing $Ker$.

Algorithm 1 is composed of 2 procedures. The first one, *initialize*, checks if $\Sigma$ is consistent. If it is the case, $Ker$ is then simply equal to $\{\Sigma\}$. Otherwise, the recursive procedure *buildKer*, which constructs $Ker$, is run.

$\Sigma$ contains the set of considered formulas and $Current$ contains the rooted consistent set we are constructing.

We use a global variable, $Ker$ for saving the final result.

The algorithm computes all rooted consistent set just one time. If $Current \cup M$ is inconsistent, we are then sure that $Current$ belongs to $Cons$, according to Proposition 20 (line 3). Then all elements containing $Current$ are removed from $Ker$ (line 4). Lastly $Current$ is added to $Ker$ (line 5).

If $Current \cup M$ is not minimal, we then choose an element $m$ of $M$ and we add it to $Current$ and we call recursively *buildKer* (line 9). When we have treated all rooted consistent subsets which contain $m$, we remove $m$ from $\Sigma$ (line 10).

Finally, with the help of proposition 18, we remove from $\Sigma$ all strictly preferred formulas to $m_i$. Indeed if $m_i$ does not belong to an element of $Ker$, then elements which are strictly preferred to $m_i$ will not belong neither.

**Example 13** *Let us apply Algorithm 1 to Example 1. As $\Sigma$ is inconsistent, we run buildKer($\{a, \neg a, b, \neg b\}$) (procedure initialization). We have $Current = \emptyset$, $\Sigma =$*

**Algorithm 1:** Algorithm for computing $Ker$

$Ker$ is a global variable which contains the final result;
**procedure** *initialization($\Sigma$);*
**begin**
　　$Ker \leftarrow \emptyset$;
　　**if** $\Sigma$ *is consistent* **then**
　　　　buildKer($\Sigma, \emptyset, \emptyset$)
　　**else**
　　　　$Ker \leftarrow \{\Sigma\}$
**end**
**procedure buildKer** ($\Sigma, \preceq_\Sigma, Current, minBack$);
**begin**
1　$M \leftarrow Min(\Sigma \setminus Current, \prec_{\Sigma \setminus Current})$ ;
2　**if** $M \neq \emptyset$ **then**
3　　**if** $Current \cup M \cup minBack$ *is inconsistent* **then**
4　　　　$Ker \leftarrow Ker \setminus \{C \in Ker : Current \subset C\}$;
5　　　　$Ker \leftarrow Ker \cup \{Current\}$;
6　　**else**
7　　　　**foreach** $m_i \in M$ **do**
8　　　　　　**if** $(\not\exists C \in Ker\ st\ Current \cup \{m \in M : m \approx_\Sigma m_i\})$ **then**
9　　　　　　　　buidKer ($\Sigma \setminus \{m \in M : m \approx_\Sigma m_i\}, \preceq_{\Sigma \setminus \{m \in M : m \approx_\Sigma m_i\}}, Current \cup \{m \in M : m \approx_\Sigma m_i\}, minBack$);
10　　　　$\Sigma \leftarrow \Sigma \setminus \{m \in M : m \approx_\Sigma m_i\}$;
11　　　　$\Sigma \leftarrow \Sigma \setminus \{m \in M : m_i \prec_\Sigma m\}$;
12　　　　$minBack \leftarrow minBack \cup \{m \in M : m \approx_\Sigma m_i\}$;
13　　　　$M \leftarrow M \setminus \{m \in M : m \approx_\Sigma m_i\}$;
**end**

*$\{a, \neg a, b, \neg b\}$ and $minBak = \emptyset$. We compute $M = Min(\Sigma \setminus Current, \prec_{\Sigma \setminus Current})$ (line 1), which is equivalent to $Min(\Sigma \setminus Current, \prec_\Sigma)$. We then have $M = \{a, b\}$. At the first step, $M$ contains minimal elements of $\Sigma$.*

*As $Current \cup M \cup minBack$ is consistent (line 3), we choose an element in $M$ (line 7). Let $a$ be such element. We call recursively buildKer (line 9). Now we have $Current = \{a\}$, $\Sigma = \{\neg a, b, \neg b\}$ and $minBak = \emptyset$. We have $M = \{b, \neg a\}$. As $Current \cup M \cup minBack$ is inconsistent, we know that $Current$ belongs to $Cons$ and can belong to $Ker$. As $Ker$ is empty, we just add $Current$ to $Cons$ (line 5).*

*We come back to the previous level of recurrence. We have treated the case of $a$. Therefore we can remove it from $\Sigma$ (line 10). As we know that $a$ will not appear in a next $Current$, we can remove all preferred formulas, namely $\neg b$ from $\Sigma$ (line 11, according to Proposition 18, elsewhere $Current$ is not rooted). But we have to remind $a$ when computing the consistency of $Current \cup M \cup minBack$. Thus $a$ into $minBack$ will contain $\{a\}$ (line 12).*

*Now we choose the next element in $M$, namely $b$, and we call recursively buildKer. We have $Current = \{b\}$, $\Sigma = \{\neg a\}$ and $minBak = \{a\}$. Therefore, we have $M = \{\neg a\}$.*

*Current* ∪ *M* ∪ *minBack* is inconsistent, we add $\{b\}$ to *Ker*, we remove $b$ and $\neg a$ from *Ker* and the algorithm stops. The final result is $Ker = \{\{a\}, \{b\}\}$.

Note that, for this example, we need just 3 tests of consistency.

## 7 Conclusions

This paper has proposed an extension of possibilistic logic to deal with partially ordered knowledge. It completes and expands results obtained in [Benferhat et al., 2003]. More precisely, the new contributions of this paper can be summarized as follows:

**(i)** We have proposed a natural justification of possibilistic inference relation based on a set of compatible totally ordered knowledge bases.
**(ii)** We have shown that the main properties of standard possibilistic logic hold for partially ordered knowledge. This is particularly true when dealing with inconsistent beliefs. Indeed, the set of plausible consequences of partially ordered knowledge base is always consistent.
**(iii)** We have provided an analysis of properties of the set of preferred consistent subbase (namely *Cons*).
**(iv)** Lastly, an algorithm for computing plausible inferences, exploiting the properties of *Cons* has been proposed.

A future work is to related our approach with the one based on plausibility measures proposed by Halpern [Halpern, 2001]. Another future work is to apply the proposed algorithm in geographical information systems where available information is often partially ordered.

## 8 acknowledgement

This work has been supported by the European project **REV!GIS** #IST-1999-14189.
Web site: *http://www.cmi.univ-mrs.fr/REVIGIS*.